\pdfoutput=1

\documentclass[11pt]{article}

\usepackage{EMNLP2023}

\usepackage{times}
\usepackage{latexsym}

\usepackage{listings}
\usepackage[T1]{fontenc}

\usepackage[utf8]{inputenc}

\usepackage{microtype}

\usepackage{zi4}
\usepackage[T1]{fontenc}

\usepackage{booktabs}
\usepackage{pifont}
\newcommand{\cmark}{\ding{51}}%
\newcommand{\xmark}{\ding{55}}%
\usepackage{graphicx}
\usepackage{amsmath}
\DeclareFontFamily{OT1}{pzc}{}
\DeclareFontShape{OT1}{pzc}{m}{it}{ <-> s*[1.2] pzcmi7t }{}
\DeclareMathAlphabet{\mathpzc}{OT1}{pzc}{m}{it}

\title{Gradually Excavating External Knowledge for Implicit Complex Question Answering}

\author{
    Chang Liu$^{1}$, Xiaoguang Li$^{2}$, Lifeng Shang$^{2}$, Xin Jiang$^{2}$, Qun Liu$^{2}$, \\ \textbf{Edmund Y. Lam$^{1}$, Ngai Wong$^{1}$} \\
    $^{1}$The University of Hong Kong, $^{2}$Huawei Noah's Ark Lab \\
    \texttt{lcon7@connect.hku.hk} \\ \texttt{\{lixiaoguang11, Shang.Lifeng, Jiang.Xin, qun.liu\}@huawei.com} \\
    \texttt{\{elam, nwong\}@eee.hku.hk}
}

\begin{document}
\maketitle
\begin{abstract}
Recently, large language models (LLMs) have gained much attention for the emergence of human-comparable capabilities and huge potential. However, for open-domain implicit question-answering problems, LLMs may not be the ultimate solution due to the reasons of: 1) uncovered or out-of-date domain knowledge, 2) one-shot generation and hence restricted comprehensiveness. To this end, this work proposes a gradual knowledge excavation framework for open-domain complex question answering, where LLMs iteratively and actively acquire external information, and then reason based on acquired historical knowledge. Specifically, during each step of the solving process, the model selects an action to execute, such as querying external knowledge or performing a single logical reasoning step, to gradually progress toward a final answer. Our method can effectively leverage plug-and-play external knowledge and dynamically adjust the strategy for solving complex questions. Evaluated on the StrategyQA dataset, our method achieves 78.17\% accuracy with less than 6\% parameters of its competitors, setting new SOTA for $\sim$10B-scale LLMs.
\end{abstract}

\section{Introduction}
\begin{figure*}[ht]
    \centering
    \includegraphics[width=\textwidth]{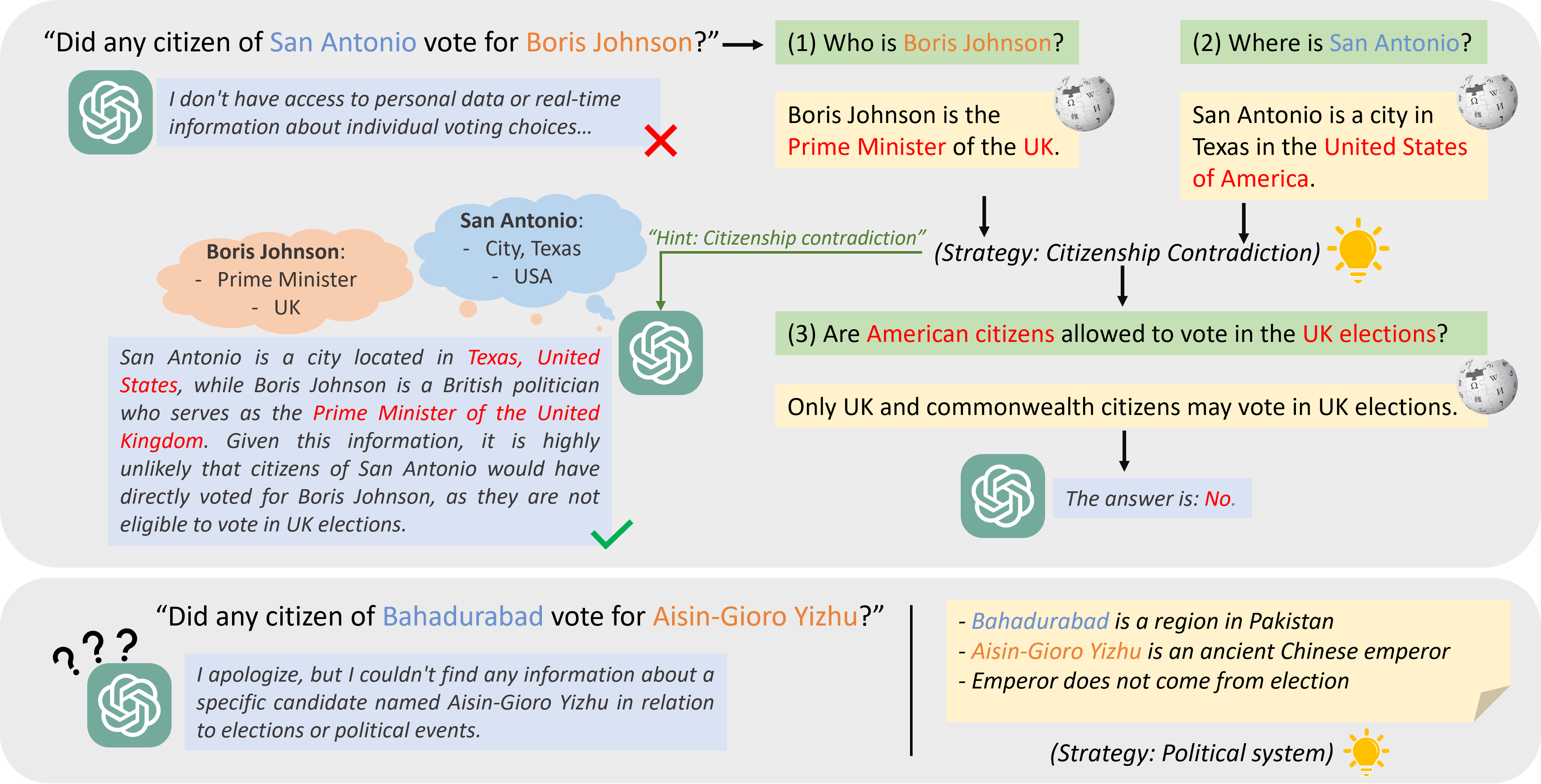}
    \caption{LLMs fail to solve open-domain complex questions due to unrecognized entities and implicit strategies. (1) In the upper part, the LLM fails to answer the question with the one-shot generation, for there is no off-the-shelf answer or evidence to this question. However, the question can be decomposed into several sub-questions and be solved once the citizenship contradiction is identified. If the hint of `citizenship contradiction' is also given, the LLM can successfully solve the question with the inner knowledge now. (2) But for the bottom case with less well-known entities, the LLM fails again due to a lack of specialized knowledge about `Aisin-Gioro Yizhu' and hence rejects to answer. Moreover, the strategy of the `political system' is not likely to be discovered from the question text only, unless enough knowledge is provided. `Citizenship contradiction' is also a possible solution.}
    \label{fig:fig1_motivation}
\end{figure*}

Recently, powerful LLMs such as ChatGPT, GPT4~\cite{openai2023gpt4}, PaLM~\cite{anil2023palm2}, LLaMA and its variances~\cite{touvron2023llama, taori2023alpaca}, exhibiting human-alike ability in conversation. It is believed the LLMs memorize knowledge in their parameters from the vast pretraining data~\cite{moiseev2022skill, roberts2020how_much_knowledge}. Nonetheless, they could still fail to solve open-domain implicit complex questions. 

In real-world applications, users might ask questions in arbitrary domain that requires specific knowledge, and expect the model to return not only syntactically fluent but also factually correct answers. Beyond open-domain, the questions can also be multi-step and implicit, consisting of multiple sub-questions that cannot be directly identified from the question language, but require logical reasoning to form a problem-solving strategy. Because of the above challenging characteristics, how to answer open-domain implicit complex questions remains an open question.

For example, in the upper part of Fig.~\ref{fig:fig1_motivation}, an implicit complex question ``Did any citizen of San Antonio vote for Boris Johnson" confuses the LLM because there is no direct information about individual voting history. However, the strategy of searching voting history is invalid does not mean the question is unsolvable. On the top right corner of Fig.~\ref{fig:fig1_motivation}, the question can be decomposed into sub-questions about Boris Johnson and San Antonio, respectively, and a strategy of checking citizenship can be easily unsealed from the background knowledge (marked in red). Following the strategy, the answer is straightforward because US citizens cannot vote in UK elections. Now if the question is re-asked with the hint of `citizenship contradiction', the LLM can successfully recall its inner knowledge about Boris Johnson and San Antonio, hence correctly answering the question. However, different from existing works that try to design manual prompts~\cite{weichain,lyu2023faithful_cot} to serve as the hint in the above example, we want to stress that this approach is not always valid. Because the key `citizenship contradiction' is not directly linked to the question text, but relies on the gradually increased knowledge during the solving process (e.g., background about Boris Johnson and San Antonio).

Another problem for solving open-domain complex questions with LLMs is the finite pretrained knowledge. In the bottom part of Fig.~\ref{fig:fig1_motivation}, we rewrite the question with two less well-known entities out of the LLM's knowledge scope, and it fails again due to limited knowledge.

Therefore, in this work, we propose a pipeline named Gradually Excavating External Knowledge (\textbf{GEEK}) to address the main challenges of open-domain implicit complex question answering: external knowledge, multi-step complexity, and implicit logic strategy. Given an open-domain multi-step implicit question, GEEK progressively decomposes the problem into several sub-questions, and iteratively calls different modules for answering the sub-questions. In the end, a final answer is concluded, synthesizing the historical sub-questions and their corresponding answers. Specifically, GEEK consists of three modules, core model, retriever, and extractor. The core model handles logical reasoning and selects an action to perform at each time step, planning the solving strategy purposefully. The retrievers allocate relevant context paragraphs from the external corpus (e.g., Wikipedia) to provide trustworthy knowledge, and the extractor condenses the textual knowledge into brief fact sentences.

During intermediate steps, GEEK can adjust the rest sub-questions based on the gradually increased external knowledge, hence forming a valid strategy like in Fig.~\ref{fig:fig1_motivation}. Considering there usually exist multiple valid solutions for a question, we enable GEEK to branch out different sub-questions during the solving process, thus exploring a strategy space and improving the final accuracy. 
 
We verified GEEK on the challenging StrategyQA dataset~\cite{geva2021strategyQA}, which consists of open-domain multi-step implicit questions. GEEK achieves 78.17\% accuracy with less than 6\% parameters of its competitors, refreshing the new SOTA for LLMs under $\sim$300B scale.

Our main contributions are threefold:
\begin{itemize}
    \item We propose GEEK, a novel pipeline to solve open-domain complex questions by progressively acquiring external knowledge and adjusting its strategy.
    \item GEEK is able to explore a strategy space to solve the question with different approaches, hence improving the overall performance.
    \item Our method is evaluated on the challenging StrategyQA benchmark, achieving 77.73\% accuracy, surpassing vanilla LLMs such as ChatGPT with 94\% parameters less.
\end{itemize}

\section{Related Work}
\textbf{\emph{Retrieving external information}} is a widely adopted method that can provide flexible knowledge to extend LLMs for specific-domain tasks. \cite{izacard2020leveraging} leverages retrievers (e.g., BM25~\cite{robertson1995bm25} or DPR~\cite{karpukhin2020dpr}) to collect relevant passages, and fuse them in the decoder for a final answer. Similarly, \cite{zhu2021adaptive} proposes AISO, which performs multi-round retrieval with different retriever models of BM25, DPR, and LINK. It then synthesizes the retrievals for a comprehensive conclusion. HopRetriever~\cite{li2021hopretriever} adopts a multi-hop manner, which identifies significant entities in previous retrievals, and then uses the entities as queries for next step retrieval.

 \

\noindent\textbf{\emph{Multi-step Implicit Question Answering}} involves complex questions that consist of several single-step questions, whose answers can be directly founded in reference context or inferred via logical deduction. The question is implicit if the decomposition strategy cannot be formulated merely from the question text. StrategyQA~\cite{geva2021strategyQA} is a dataset for multi-step implicit question answering, including human-annotated solutions in the form of decomposition questions and corresponding fact sentences from Wikipedia. However, while human beings can achieve 87\% accuracy~\cite{geva2021strategyQA}, the dataset has proved to be very challenging for language models to solve. Merely improving the quality of retrieval~\cite{liang2022betterRetrieval} or decomposition strategy~\cite{katz2022inferringImplicitRelations} cannot effectively boost final answer accuracy.

Several previous studies adopt the design of iterative retrieval and reasoning to solve multi-step complex questions. IRGR~\cite{ribeiro2022entailment_iterativeRetrieval} performs iteratively retrieval to search for suitable premises. ReAct~\cite{yao2022react} lets the model first generate a reasoning sentence about what next action (e.g., search via web API) to be performed, and then execute the selected action. Maieutic Prompting~\cite{jung2022maieutic} introduces a maieutic tree that recursively entails component statements, and uses the concept of `logical integrity' to verify each step. RR~\cite{he2022rethinking} combines CoT~\cite{weichain} with retrieval to verify the correctness of the reasoning process, and achieved an accuracy of 77.73\% on the StrategyQA dataset, the SOTA for LLMs below $\sim$300B scale at that time.

Nonetheless, most of the previous studies using iteratively reasoning for multi-step question answering only focus on datasets such as HotpotQA~\cite{yang2018hotpotqa}, which do not involve implicit questions.  Instead, their method assumes a straightforward direction from one step to another. For example, IRCoT~\cite{trivedi2022interleaving_retrieval} directly uses the `thought' (intermediate result) from the last step as the query for next-step retrieval, which highly relies on the strong connections between the entities from each step. Different from the above-listed methods, our GEEK modeling the process of purposely excavating external knowledge by composing sub-questions, and formulating a complete strategy gradually during the solving process. 

\section{Problem definition}
In this work, we focus on open-domain implicit complex question answering. Given a question ${q}$, the model is asked to derive the final answer $\mathpzc{z}$. Specifically, ${q}$ is open-domain and hence requires some certain background facts to solve, denoted as $\mathcal{F}=\{{f}_i\}$. The facts come from an external corpus $\mathcal{C}$ (e.g., Wikipedia), not necessarily included in the model's pretraining dataset. The question ${q}$ is also multi-step, meaning that it can be decomposed into several decomposition questions $\mathcal{D}=\{{d}_i\}$. Each ${d}_i$ corresponds to a background fact ${f}_i$. Noted that $\mathcal{D}$ is usually implicit from ${q}$, which means some ${d}_i$ can only be formulated until enough facts $\mathcal{F}_i \subset \mathcal{F}$ is uncovered. Under the GEEK scenario, we define the question state $\mathcal{Q}$ at step $t$ as:
$$\mathcal{Q}_t=(q, \{(d_1, f_1), ..., (d_{t-1}, f_{t-1})\} [,(d_t, f_t)])$$,
where the current decomposition $d_t$ and fact $f_t$ may or may not have been decided yet. $\mathcal{Q}$ includes the question $q$ and historical exploration steps.

\section{Gradually Excavating External Knowledge}
In this section, we introduce GEEK for open-domain complex question answering. GEEK consists of three components, the core model, retriever and extractor model. Iteratively, the core model selects actions to perform from an action space $\mathcal{A}$, conditioned on the current question state $\mathcal{Q}$. Then the selected action is executed and the question state is updated, gradually accumulating external knowledge until a final answer can be drawn.

\begin{figure*}[ht]
    \centering
    \includegraphics[width=0.98\textwidth]{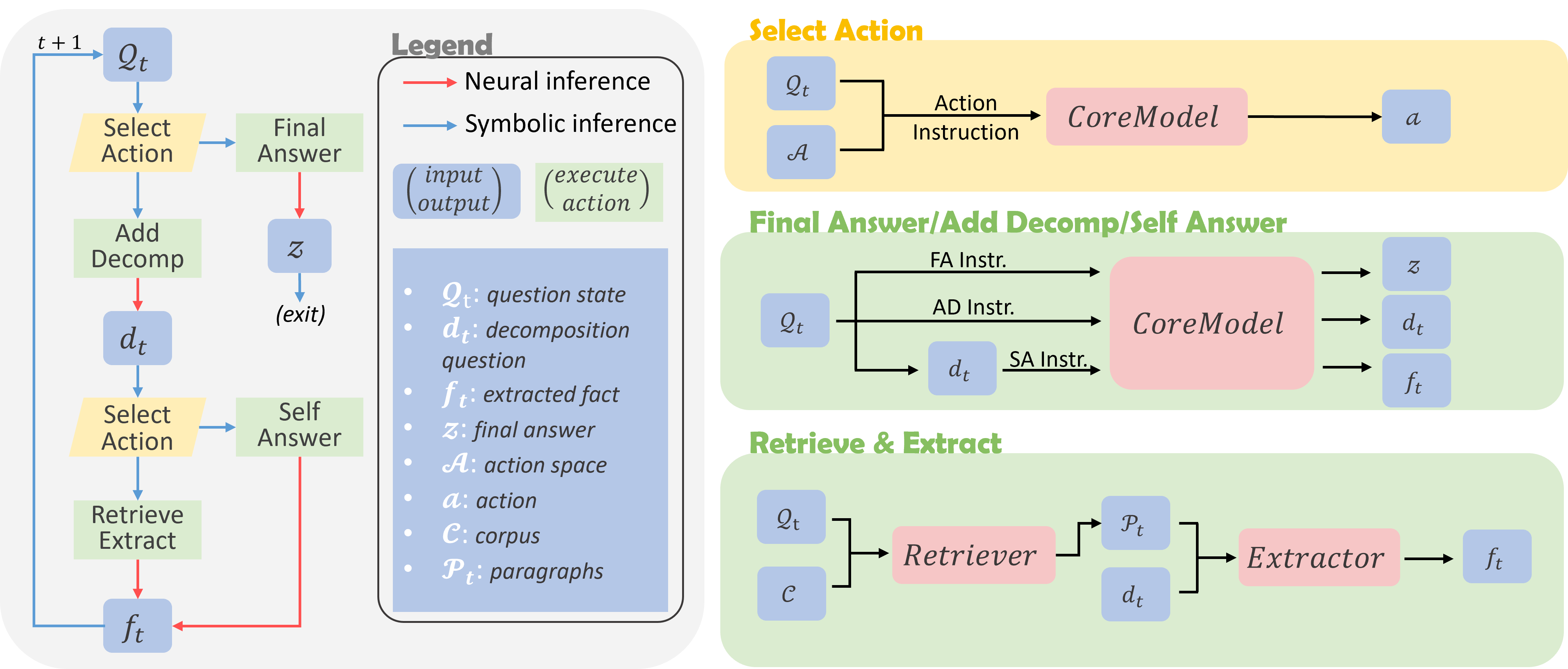}
    \caption{GEEK workflow: the core model, retriever and extractor collaborate to solve complex questions progressively. (Left): In each iteration, based on the question state $\mathcal{Q}_t$, GEEK selects an action and calls the corresponding module to execute. The execution updates the question state in turn, until a final answer $\mathpzc{z}$ is derived. (Right): The detailed procedure of action selection and execution. For action selection, $\mathcal{Q}_t$ and $\mathcal{A}$ are fed into the core model with the instruction for action selection, and the model outputs an action code $\mathpzc{a}$. For the execution of Add Decomp, Final Answer, and Self Answer, the core model outputs corresponding responses following different instructions. At last, for Retrieve \& Extract, the retriever firstly retrieves several paragraphs $\mathcal{P}$ from the corpus as external knowledge, and then the extractor answers the decomposition question $d_t$ based on $\mathcal{P}$.}
    \vspace{-5pt}
    \label{fig:fig2-workflow}
\end{figure*}

\subsection{Core model}
\label{sec:core_model}
The core model is a pretrained LLM for sequence-to-sequence text generation (e.g., Flan-T5~\cite{chung2022flan}), acting as the controller of GEEK. For each step $t$, the core model chooses an action $a$ to perform, among the action space $\mathcal{A}$ and conditioned on the current question state $\mathcal{Q}_t$:
$$a = Core(Q_t, A)$$
Besides choosing actions, the core model is also used to execute some types of actions, under action-specific prompts, such as generating decomposition questions. Details are in Sec.~\ref{sec:action_space}.

\subsection{Retriever}
In order to utilize external knowledge, we employ a neural retriever, DPR~\cite{karpukhin2020dpr}, to retrieve paragraphs from a vast volume of context $\mathcal{C}$. Given the decomposition question $d_t$ as query, the retriever returns top $k$ relevant paragraphs:
$$\mathcal{P}_t=\{p_1, p_2, ..., p_k\}=Retr(d_t, \mathcal{C})$$,
where $Retr$ stands for the retriever and each $p_i$ denotes a paragraph. Due to the enormous amount of context in $\mathcal{C}$, full-size retrieval is time costly. For efficiency, we use two nested DPR bi-encoder models~\cite{karpukhin2020dpr}, namely the document retriever and the paragraph retriever. Firstly, we deploy the title retriever for a document-level retrieval, to shrink the context space to $k_{doc}=100$ documents, where the context embeddings are built from each document's title and first paragraph. Then the paragraph retriever performs a secondary search on the paragraph level, outputting the top $k$ matched paragraphs among the 100 documents. 

\subsection{Extractor}
Even though we retrieve $k$ paragraphs from the vast $\mathcal{C}$, these retrieved paragraphs are still too long to be input into the core model. Therefore, another specialized extractor is used to condense the $k$ paragraphs into concise fact sentences for $d_t$:
$$f_t=Extractor(d_t, \mathcal{P})$$
We use FiD architecture~\cite{izacard2020fid} for the extractor. Instead of extracting facts locally from each of the $k$ paragraphs, FiD can perceive all the paragraphs simultaneously and generate more comprehensive results.

\subsection{GEEK Pipeline and Action Space}
\label{sec:action_space}

As shown in Fig.~\ref{fig:fig2-workflow}, GEEK iteratively selects and executes actions to solve implicit complex questions. The procedure involves neural inference where neural networks generate text outputs that are less interpretable due to the black-box nature of neural networks, as well as symbolic inference which follows strict rules. Based on the question state $\mathcal{Q}_t$ at each round, the core model decides whether a final answer can be made or more decomposition needs to be explored. If the latter, it also determines whether external knowledge should be retrieved, or the decomposition can be directly answered using the fact sentences so far. The newly acquired facts are added to the question state for the next iteration. Next, we introduce the details for each action:

\begin{itemize}
    \item \textbf{FinalAnswer} (\emph{core model}) If enough background knowledge is acquired and a final conclusion can be drawn, the core model should output the final answer to $q$. For the real implementation, we let the model summarize the facts from previous steps as a self-CoT, and then conclude the final answer (yes or no).

    \item \textbf{AddDecomp} (\emph{core model}) Based on current $\mathcal{Q}_{t}$, the core model generate a next-step decomposition question $d_{t}$. The generation is conditioned on previous decompositions and facts. Hence the model could adjust the reasoning strategy on-fly, and benefit from the gradually enhanced external knowledge.
    
    To further improve the comprehensiveness of the strategy and avoid generating unsolvable decomposition questions, we design a \emph{pre-answer} trick. Instead of generating $d_t$ only, we use an explicit prompt to instruct the model to generate all the remaining decompositions with corresponding pseudo answers: $(d_t, \tilde{f}_t, d_{t+1}, \tilde{f}_{t+1}, \cdots)$. Hence, the strategy leading by $d_t$ could be more coherent and solvable. Note that the generated pre-answers are not necessarily correct, and all the redundant generations except $d_t$ only serve as generation auxiliaries, which will be removed before adding to the question state $\mathcal{Q}_t$.

    \item \textbf{Retrieve \& Extract} (\emph{retriever}) Once a new decomposition is added, the core model would decide whether this decomposition question needs external knowledge to answer. If yes, this action is executed and the retriever is called to retrieve the top $k$ most relevant paragraphs $\mathcal{P}_t$ corresponding to the current decomposition $d_{t}$. Once $\mathcal{P}_t$ is retrieved, the extractor model would read the $k$ paragraphs and generate a concise fact $f_t$ as the answer to $d_t$. 

    For the extractor, we also use the generated pseudo answer $\tilde{f}_t$ from the `AddDecomp' action as a reference. We formulate the input to extractor with the prompt: `Answer the question {$d_t$} based on the context {$p_i^{(t)}$}, a reference but not necessarily correct answer is {$\tilde{f}_t$}'. Therefore, the extractor knows what kind of information should be extracted among possibly multiple aspects, avoiding correct but not relevant answers (e.g., for decomposition `Who is xxx?', possible aspects include nationality, career, education, family, etc.).  

    \item \textbf{SelfAnswer} (\emph{core model}) For some decomposition questions that are pure logical deduction, or the required knowledge has been included in the question state already, no external evidence is needed and hence retriever and extractor are not used. In this case, a self-answer prompt is used and the core model would answer $d_{t}$ directly, outputting $f_{t}$.

\end{itemize}

\subsection{Strategy Exploration}
Considering that there often exist multiple possible strategies to solve an identical question, we also extend the GEEK to explore a strategy space for different solutions. Specifically, in the step of \emph{AddDecomp}, the core model can return multiple decomposition questions using beam search. For each different decomposition $d_t^{(i)}$, a copy of current question state $\mathcal{Q}_t$ is created, and updated by the new decomposition, forming $\mathcal{Q}_t^{(i)}$. Then the copies carry on for the rest solving process independently. 

We emphasize that our method Strategy Exploration (SE) is different from Self-Consistency~\cite{wang2022self_consistency}, which outputs multiple CoT solutions with the one-shot generation, and then takes the majority. Under the scenario of GEEK with SE, the question branches into $n=4$ different strategies at every iteration, hence formulating a latent solution tree. The diverged decomposition would lead to different retrieval results and generated facts, hence is an exploration of the strategy space. Due to computation constraints, we limit the expansion number to be at most 16 (i.e., expansion rate $n=4$ at each decomposition and for at most 2 iterations). The majority vote is used to derive the final answer.

\begin{figure*}[ht]
    \centering
    \includegraphics[width=0.9\textwidth]{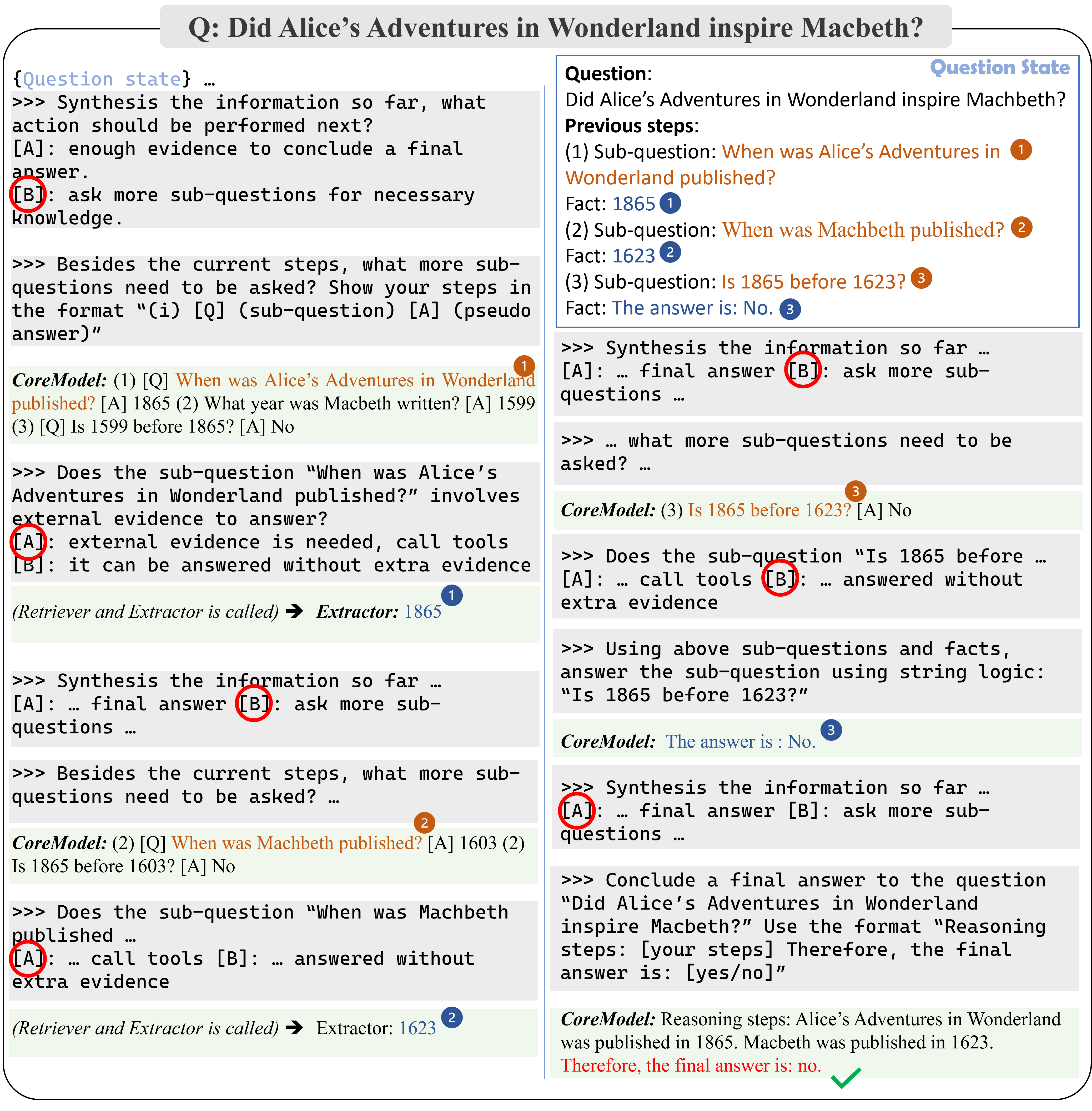}
    \caption{Full process of GEEK inference. For each round, the prompts are shown in gray, and the current question state is also given to the model as input. Model responses are shown in green and action selection is represented by a red circle to save space. On the top right corner, the question state is listed, where the historical states of sub-question and fact are gradually added during the inference (best viewed in color and numerical marks).}
    \vspace{-5pt}
    \label{fig:fig3-example}    
\end{figure*}

\section{Case Study}
In Fig.~\ref{fig:fig3-example} we show a full inference process of GEEK. The inference involves in total three iterations. In order to save space, we omit the input question state for each step, but show it in the top right corner. Marked in colored numbers, the question state is gradually enriched as the decomposition questions and corresponding facts are added. Also, for each action selection step, we circle out the core model's choice. Some repeated prompts are also abbreviated, due to the page limitation.

As can be observed in the example, the GEEK decomposes the original question into three sub-questions, and takes the strategy of `temporal confliction' to solve the problem. For sub-questions `(1)' and `(2)', external knowledge is retrieved and the original pseudo answers are verified and corrected (e.g., `1599' $\rightarrow$ `1623'). The strategy is also dynamically adjusted as more facts are acquired (e.g., `Is 1599 before 1865?' $\rightarrow$ `Is 1865 before 1623?'). For the last logical sub-question that requires no external knowledge, the core model chooses to self-answer it without calling retriever and extractor. When there is enough information, the final answer `no' is given with reasons.

\section{Experiment}
\vspace{-5pt}
In this section, we compare GEEK with previous baselines and conduct ablation studies to analyze the contributions of specific modules.

\begin{table*}[ht]
    \centering
    \begin{tabular}{llccc}
    \toprule
    Method & Backbone & Retrieve & Specification & SQA \\
    \midrule
    ChatGPT~\cite{qin2023is_chatgpt_a_general_solver} & GPT-3.5 (175B) & \xmark & 
 Without CoT& 59.2 \\
    ChatGPT~\cite{qin2023is_chatgpt_a_general_solver} & GPT-3.5 (175B) & \xmark & CoT & 62.5 \\
    FaithfulCoT~\cite{lyu2023faithful_cot} & 
    code-davinci-002 (175B) & \xmark  & - & 73.2 \\
    \cite{xie2023decompositionEnhances} & 
    code-davinci-002 (175B) & \xmark & - & 77.2 \\
    \cite{lazaridou2022internet_augmented_lm} & Gopher (280B) & \cmark & - & 66.2 \\
    Visconde~\cite{pereira2023visconde} & text-davinci-002 (175B) & \cmark & CoT & 69.43 \\
    RR~\cite{he2022rethinking} & 
    text-davinci-002 (175B) & \cmark & CoT & \textit{77.73} \\ 
    \midrule
    PaLM~\cite{chowdhery2022palm} & PaLM (540B) & \xmark & - & 73.9 \\
    PaLM~\cite{anil2023palm2} & PaLM (540B) & \xmark & CoT + SC & 81.6 \\
    PaLM2~\cite{anil2023palm2}  & PaLM2 (340B) & \xmark & - & \textbf{90.4} \\
    \midrule
    GEEK (ours) & Flan-T5 (11B) & \cmark & CoT & 75.98 \\
    GEEK (ours) & Flan-T5 (11B) & \cmark & CoT+SE & \textbf{78.17} \\
    \bottomrule
    \end{tabular}
    \caption{Experiment results on strategyQA dataset. GEEK achieves the SOTA accuracy for LLMs in $\sim$10B scale, and surpasses all the previous methods with backbone under 300B scale, using only 6\% parameters or less.}
    \vspace{-5pt}
    \label{tab:main_results}
\end{table*}

\subsection{Dataset and Preprocessing}
We use the StrategyQA dataset~\cite{geva2021strategyQA} to evaluate our method. The dataset consists of 2061 samples in the train set and 229 in the dev set. Another 490 samples without labels are provided for the test set, and results for that can be uploaded to their website for evaluation~\footnote{https://allenai.org/data/strategyqa}. For train and dev samples, each question is provided a human-annotated strategy $\mathcal{D}$ in the form of decomposition questions. Golden supporting paragraphs $p$ for each $d$ are also given, which are from 36.6M Wikipedia processed corpus. In addition, human-annotated background facts $\mathcal{F}$ are also provided.

Nonetheless, the provided background facts are not strictly mapped with the decomposition, either in the aspect of total number or sequential order. Additionally, over 25\% human annotated decomposition questions refer to previous $i$-th decomposition by symbol `\#i'. We find that these symbols cannot be simply filled by decomposition's corresponding answer, because some `\#i' may refer to an entity in the decomposition question text. For the above two reasons, we use GPT4~\cite{openai2023gpt4} to refine the annotations. We provide the question together with the final answer, golden facts and decomposition questions to GPT4, and prompt it to fill the '\#i's in annotations. Meanwhile, we also ask it to give a concise answer for each decomposition, according to the golden facts and the final answer. Unless specified, all the experiment results below are from the GPT4 processed version.

\subsection{Detailed Settings}
For retriever, vanilla DPR~\cite{karpukhin2020dpr} is used with BERT-base-uncased~\cite{devlin2018bert} as backbone. By default, $k=10$ paragraphs will be retrieved during GEEK inference. The extractor is a FiD~\cite{izacard2020fid} with Flan-T5-3B as the backbone. During training, the extractor is fed golden paragraphs and remaining retrieval paragraphs, to satisfy the fixed number $k$.

For the core model, we adopt Flan-T5-11B~\cite{chung2022flan, raffel2020exploring_T5}. The model is trained for multiple tasks including action selection and execution of three actions, as described in Fig.~\ref{fig:fig2-workflow}. The tasks are trained in parallel, with input-output pairs built from human annotations and in the same instruction format as Fig.~\ref{fig:fig3-example}. We train the model on 8 V100 GPUs. Due to the LLMs' out-of-memory problems, the deepspeed flatform is used~\cite{rasley2020deepspeed}. During inference, we utilize the accelerate package with offloading for acceleration~\cite{accelerate}.

\subsection{Comparison with other Baselines}
In this section, we compare the proposed GEEK with other baselines on the strategyQA dataset. As shown in Tab.~\ref{tab:main_results}, GEEK yields 78.17\% accuracy on the StrategyQA dataset, with a much smaller model size (11B) than previous baselines. Our method sets a new SOTA for the $\sim$10B LLMs, and also is the second best among all existing methods except PaLM. Although GEEK is finetuned with supervision, the accuracy of GEEK is still considerably high considering its size in comparison with other baselines. It is also worth noting that due to the indirection of using LLM APIs, finetuning or adaption for special domain tasks is not easy. For example, only Visconde~\cite{pereira2023visconde} and RR~\cite{he2022rethinking} successfully leverage external knowledge, but with a relatively less capable backbone text-davinci-002. However, the external knowledge is proved to benefit the task effectively, and RR achieves SOTA performance at its time. 

Specifically, while the initially finetuned GEEK yields 75.98\% accuracy, by using SE, the accuracy is improved significantly, to 78.17\%. This process requires no retraining yet can boost final accuracy, where GEEK explores the strategy space and tries to solve the question via multiple paths.

\subsection{Ablation Study}
We also analyze the contribution of the different components in GEEK. Results are shown in Tab.~\ref{tab:abaltion_results}. The line of `CoT' denotes that the core model directly answers the question as in the `FinalAnswer' action, following the CoT approach~\cite{wei2022chain} without iterative reasoning and knowledge retrieval. After finetuning, the accuracy is 70.74\% accuracy, 8.73\% higher than zero-shot Flan-T5 but 7.73\% lower than the full version of GEEK. We find that performing the action `Retrieve\&Extract' could efficiently increase the accuracy to 75.98\%, justifying the motivation of leveraging external knowledge for solving open-domain questions.

\begin{table}[t]
    \centering
    \begin{tabular}{lccccc}
    \toprule
     & De & RE & SE & Acc \\
     \midrule
    Zero-shot & \xmark & \xmark & \xmark & 62.01 \\
    CoT & \xmark & \xmark & \xmark & 70.74 \\
    +De & \cmark & \xmark & \xmark & 71.50\\
    +RE & \cmark & \cmark & \xmark & 75.98 \\
    Full & \cmark & \cmark & \cmark & 78.17 \\
    \bottomrule
    \end{tabular}
    \caption{Ablation study results. The three columns denote whether an action is performed under GEEK. (`De': `AddDecomp', `RE': `Retrieve and Extract', `SE': `Strategy Exploration')}
    \label{tab:abaltion_results}
\end{table}

\begin{table}[t]
    \centering
    \begin{tabular}{lcccc}
    \toprule
     & Human & Equal & GEEK  \\
     \midrule
    ChatGPT & 13.54\% & 24.02\% & 62.45\% \\ 
    \bottomrule
    \end{tabular}
    \caption{ChatGPT Assessment of prediction results.}
    \vspace{-5pt}
    \label{tab:chatgpt_results}
\end{table}

\subsection{ChatGPT Assessment}
To further evaluate the quality of the strategy generated by GEEK, we also leverage the GPT4 to simulate a human assessment. Specifically, we show it the human-annotated decomposition questions and corresponding facts, as well as the GEEK-generated decomposition-fact pairs, and let the model choose which is more informative and faithfully correct. The results are listed in Tab.~\ref{tab:chatgpt_results}. It turns out that 62.45\% of the GEET-generated decomposition-fact pairs are preferred.

\section{Conclusion}
We present GEEK, a pipeline to progressively excavate external knowledge for boosting LLM's capability in solving open-domain multi-step implicit questions. Interactively, GEEK decomposes the question into explicit sub-questions to retrieve external knowledge, and the accumulated knowledge enlightens the model to form a better strategy in turn. The proposed method also provides explainability by showing the full reasoning process, supported by retrieved evidence. By using SE, GEEK can also explore the strategy space and try different approaches to solve the complex question, which increases the performance further. Experiment results justify our design. With GEEK, 78.17\% accuracy is achieved using less than 6\% size of the competitors, refreshing the SOTA accuracy for $\sim$10B LLMs. As an alternative to the paradigm of scaling for larger models and more pretraining data, we hope this research could inspire more future works to investigate how to organically excavate external knowledge and progressively formulate strategies for solving open-domain implicit questions.

\section*{Limitations}
Albeit achieving outstanding overall accuracy with a significantly smaller model size, the GEEK is not without limitations. First of all, as long as neural reasoning is involved, the hallucination problem is inevitable in theory, due to the black-box nature of neural networks. We want to stress that by using a retriever and an extractor, the hallucination problem can be alleviated from factual references, but not completely avoided. Secondly, the logic of GEEK is not guaranteed to be correct. It is possible that GEEK gives the correct answer but wrong solving steps, or correct intermediate steps but wrong final answer. Lastly, due to the scarcity of public datasets like StrategyQA for open-domain complex question answering, it is difficult for us to research this problem under more datasets and different task settings. We expect future works that may come up with more suitable datasets like StrategyQA.

\bibliography{anthology,custom}

@article{openai2023gpt4,
    title={GPT-4 technical report},
    author={OpenAI},
    year={2023},
    url={https://arxiv.org/abs/2303.08774},
    journal={Computation and Language},
    volume={arXiv:2303.08774}
}

@article{touvron2023llama,
  title={Llama: Open and efficient foundation language models},
  author={Touvron, Hugo and Lavril, Thibaut and Izacard, Gautier and Martinet, Xavier and Lachaux, Marie-Anne and Lacroix, Timoth{\'e}e and Rozi{\`e}re, Baptiste and Goyal, Naman and Hambro, Eric and Azhar, Faisal and others},
  journal={arXiv preprint arXiv:2302.13971},
  year={2023}
}

@article{taori2023alpaca,
  title={Alpaca: A strong, replicable instruction-following model},
  author={Taori, Rohan and Gulrajani, Ishaan and Zhang, Tianyi and Dubois, Yann and Li, Xuechen and Guestrin, Carlos and Liang, Percy and Hashimoto, Tatsunori B},
  journal={Stanford Center for Research on Foundation Models. https://crfm. stanford. edu/2023/03/13/alpaca. html},
  volume={3},
  number={6},
  pages={7},
  year={2023}
}

@article{anil2023palm2,
  title={Palm 2 technical report},
  author={Anil, Rohan and Dai, Andrew M and Firat, Orhan and Johnson, Melvin and Lepikhin, Dmitry and Passos, Alexandre and Shakeri, Siamak and Taropa, Emanuel and Bailey, Paige and Chen, Zhifeng and others},
  journal={arXiv preprint arXiv:2305.10403},
  year={2023}
}

@article{he2022rethinking,
  title={Rethinking with Retrieval: Faithful Large Language Model Inference},
  author={He, Hangfeng and Zhang, Hongming and Roth, Dan},
  journal={arXiv preprint arXiv:2301.00303},
  year={2022}
}

@article{geva2021strategyQA,
  title={Did aristotle use a laptop? a question answering benchmark with implicit reasoning strategies},
  author={Geva, Mor and Khashabi, Daniel and Segal, Elad and Khot, Tushar and Roth, Dan and Berant, Jonathan},
  journal={Transactions of the Association for Computational Linguistics},
  volume={9},
  pages={346--361},
  year={2021},
  publisher={MIT Press}
}

@article{chung2022flan,
  title={Scaling instruction-finetuned language models},
  author={Chung, Hyung Won and Hou, Le and Longpre, Shayne and Zoph, Barret and Tay, Yi and Fedus, William and Li, Eric and Wang, Xuezhi and Dehghani, Mostafa and Brahma, Siddhartha and others},
  journal={arXiv preprint arXiv:2210.11416},
  year={2022}
}

@article{karpukhin2020dpr,
  title={Dense passage retrieval for open-domain question answering},
  author={Karpukhin, Vladimir and O{\u{g}}uz, Barlas and Min, Sewon and Lewis, Patrick and Wu, Ledell and Edunov, Sergey and Chen, Danqi and Yih, Wen-tau},
  journal={arXiv preprint arXiv:2004.04906},
  year={2020}
}

@article{izacard2020fid,
  title={Leveraging passage retrieval with generative models for open domain question answering},
  author={Izacard, Gautier and Grave, Edouard},
  journal={arXiv preprint arXiv:2007.01282},
  year={2020}
}

@article{devlin2018bert,
  title={Bert: Pre-training of deep bidirectional transformers for language understanding},
  author={Devlin, Jacob and Chang, Ming-Wei and Lee, Kenton and Toutanova, Kristina},
  journal={arXiv preprint arXiv:1810.04805},
  year={2018}
}

@inproceedings{rasley2020deepspeed,
  title={Deepspeed: System optimizations enable training deep learning models with over 100 billion parameters},
  author={Rasley, Jeff and Rajbhandari, Samyam and Ruwase, Olatunji and He, Yuxiong},
  booktitle={Proceedings of the 26th ACM SIGKDD International Conference on Knowledge Discovery \& Data Mining},
  pages={3505--3506},
  year={2020}
}

@Misc{accelerate,
  title={Accelerate: Training and inference at scale made simple, efficient and adaptable.},
  author={Sylvain, Gugger and Lysandre, Debut and Thomas, Wolf and Philipp, Schmid and Zachary, Mueller and Sourab, Mangrulkar},
  howpublished = {\url{https://github.com/huggingface/accelerate}},
  year =         {2022}
}

@article{xie2023decompositionEnhances,
  title={Decomposition enhances reasoning via self-evaluation guided decoding},
  author={Xie, Yuxi and Kawaguchi, Kenji and Zhao, Yiran and Zhao, Xu and Kan, Min-Yen and He, Junxian and Xie, Qizhe},
  journal={arXiv preprint arXiv:2305.00633},
  year={2023}
}

@article{chowdhery2022palm,
  title={Palm: Scaling language modeling with pathways},
  author={Chowdhery, Aakanksha and Narang, Sharan and Devlin, Jacob and Bosma, Maarten and Mishra, Gaurav and Roberts, Adam and Barham, Paul and Chung, Hyung Won and Sutton, Charles and Gehrmann, Sebastian and others},
  journal={arXiv preprint arXiv:2204.02311},
  year={2022}
}

@article{raffel2020exploring_T5,
  title={Exploring the limits of transfer learning with a unified text-to-text transformer},
  author={Raffel, Colin and Shazeer, Noam and Roberts, Adam and Lee, Katherine and Narang, Sharan and Matena, Michael and Zhou, Yanqi and Li, Wei and Liu, Peter J},
  journal={The Journal of Machine Learning Research},
  volume={21},
  number={1},
  pages={5485--5551},
  year={2020},
  publisher={JMLRORG}
}

@article{zhu2021adaptive,
  title={Adaptive information seeking for open-domain question answering},
  author={Zhu, Yunchang and Pang, Liang and Lan, Yanyan and Shen, Huawei and Cheng, Xueqi},
  journal={arXiv preprint arXiv:2109.06747},
  year={2021}
}

@article{izacard2020leveraging,
  title={Leveraging passage retrieval with generative models for open domain question answering},
  author={Izacard, Gautier and Grave, Edouard},
  journal={arXiv preprint arXiv:2007.01282},
  year={2020}
}

@article{robertson1995bm25,
  title={Okapi at TREC-3},
  author={Robertson, Stephen E and Walker, Steve and Jones, Susan and Hancock-Beaulieu, Micheline M and Gatford, Mike and others},
  journal={Nist Special Publication Sp},
  volume={109},
  pages={109},
  year={1995},
  publisher={National Instiute of Standards \& Technology}
}

@article{qin2023is_chatgpt_a_general_solver,
  title={Is ChatGPT a general-purpose natural language processing task solver?},
  author={Qin, Chengwei and Zhang, Aston and Zhang, Zhuosheng and Chen, Jiaao and Yasunaga, Michihiro and Yang, Diyi},
  journal={arXiv preprint arXiv:2302.06476},
  year={2023}
}

@article{lyu2023faithful_cot,
  title={Faithful chain-of-thought reasoning},
  author={Lyu, Qing and Havaldar, Shreya and Stein, Adam and Zhang, Li and Rao, Delip and Wong, Eric and Apidianaki, Marianna and Callison-Burch, Chris},
  journal={arXiv preprint arXiv:2301.13379},
  year={2023}
}

@inproceedings{li2021hopretriever,
  title={Hopretriever: Retrieve hops over wikipedia to answer complex questions},
  author={Li, Shaobo and Li, Xiaoguang and Shang, Lifeng and Jiang, Xin and Liu, Qun and Sun, Chengjie and Ji, Zhenzhou and Liu, Bingquan},
  booktitle={Proceedings of the AAAI Conference on Artificial Intelligence},
  volume={35},
  pages={13279--13287},
  year={2021}
}

@article{liang2022betterRetrieval,
  title={Better Retrieval May Not Lead to Better Question Answering},
  author={Liang, Zhengzhong and Khot, Tushar and Bethard, Steven and Surdeanu, Mihai and Sabharwal, Ashish},
  journal={arXiv preprint arXiv:2205.03685},
  year={2022}
}

@inproceedings{katz2022inferringImplicitRelations,
  title={Inferring implicit relations in complex questions with language models},
  author={Katz, Uri and Geva, Mor and Berant, Jonathan},
  booktitle={Findings of the Association for Computational Linguistics: EMNLP 2022},
  pages={2548--2566},
  year={2022}
}

@article{ribeiro2022entailment_iterativeRetrieval,
  title={Entailment tree explanations via iterative retrieval-generation reasoner},
  author={Ribeiro, Danilo and Wang, Shen and Ma, Xiaofei and Dong, Rui and Wei, Xiaokai and Zhu, Henry and Chen, Xinchi and Huang, Zhiheng and Xu, Peng and Arnold, Andrew and others},
  journal={arXiv preprint arXiv:2205.09224},
  year={2022}
}

@inproceedings{yao2022react,
  title={ReAct: Synergizing Reasoning and Acting in Language Models},
  author={Yao, Shunyu and Zhao, Jeffrey and Yu, Dian and Shafran, Izhak and Narasimhan, Karthik R and Cao, Yuan},
  booktitle={NeurIPS 2022 Foundation Models for Decision Making Workshop},
  year={2022}
}

@article{jung2022maieutic,
  title={Maieutic prompting: Logically consistent reasoning with recursive explanations},
  author={Jung, Jaehun and Qin, Lianhui and Welleck, Sean and Brahman, Faeze and Bhagavatula, Chandra and Bras, Ronan Le and Choi, Yejin},
  journal={arXiv preprint arXiv:2205.11822},
  year={2022}
}

@inproceedings{weichain,
  title={Chain-of-Thought Prompting Elicits Reasoning in Large Language Models},
  author={Wei, Jason and Wang, Xuezhi and Schuurmans, Dale and Bosma, Maarten and Xia, Fei and Chi, Ed H and Le, Quoc V and Zhou, Denny and others},
  booktitle={Advances in Neural Information Processing Systems},
  year={2022}
}

@inproceedings{pereira2023visconde,
  title={Visconde: Multi-document QA with GPT-3 and Neural Reranking},
  author={Pereira, Jayr and Fidalgo, Robson and Lotufo, Roberto and Nogueira, Rodrigo},
  booktitle={European Conference on Information Retrieval},
  pages={534--543},
  year={2023},
  organization={Springer}
}

@article{lazaridou2022internet_augmented_lm,
  title={Internet-augmented language models through few-shot prompting for open-domain question answering},
  author={Lazaridou, Angeliki and Gribovskaya, Elena and Stokowiec, Wojciech and Grigorev, Nikolai},
  journal={arXiv preprint arXiv:2203.05115},
  year={2022}
}

@article{roberts2020how_much_knowledge,
  title={How much knowledge can you pack into the parameters of a language model?},
  author={Roberts, Adam and Raffel, Colin and Shazeer, Noam},
  journal={arXiv preprint arXiv:2002.08910},
  year={2020}
}

@inproceedings{moiseev2022skill,
  title={SKILL: Structured Knowledge Infusion for Large Language Models},
  author={Moiseev, Fedor and Dong, Zhe and Alfonseca, Enrique and Jaggi, Martin},
  booktitle={Proceedings of the 2022 Conference of the North American Chapter of the Association for Computational Linguistics: Human Language Technologies},
  pages={1581--1588},
  year={2022}
}

@article{wang2022self_consistency,
  title={Self-consistency improves chain of thought reasoning in language models},
  author={Wang, Xuezhi and Wei, Jason and Schuurmans, Dale and Le, Quoc and Chi, Ed and Narang, Sharan and Chowdhery, Aakanksha and Zhou, Denny},
  journal={arXiv preprint arXiv:2203.11171},
  year={2022}
}

@inproceedings{yang2018hotpotqa,
  title={HotpotQA: A Dataset for Diverse, Explainable Multi-hop Question Answering},
  author={Yang, Zhilin and Qi, Peng and Zhang, Saizheng and Bengio, Yoshua and Cohen, William and Salakhutdinov, Ruslan and Manning, Christopher D},
  booktitle={Proceedings of the 2018 Conference on Empirical Methods in Natural Language Processing},
  pages={2369--2380},
  year={2018}
}

@article{trivedi2022interleaving_retrieval,
  title={Interleaving Retrieval with Chain-of-Thought Reasoning for Knowledge-Intensive Multi-Step Questions},
  author={Trivedi, Harsh and Balasubramanian, Niranjan and Khot, Tushar and Sabharwal, Ashish},
  journal={arXiv preprint arXiv:2212.10509},
  year={2022}
}

@article{wei2022chain,
  title={Chain-of-thought prompting elicits reasoning in large language models},
  author={Wei, Jason and Wang, Xuezhi and Schuurmans, Dale and Bosma, Maarten and Xia, Fei and Chi, Ed and Le, Quoc V and Zhou, Denny and others},
  journal={Advances in Neural Information Processing Systems},
  volume={35},
  pages={24824--24837},
  year={2022}
}

@article{hartill2023answering,
  title={Answering Unseen Questions With Smaller Language$\backslash$$\backslash$Models Using Rationale Generation and Dense Retrieval},
  author={Hartill, Tim and Benavides-Prado, Diana and Witbrock, Michael and Riddle, Patricia J},
  journal={arXiv preprint arXiv:2308.04711},
  year={2023}
}

@article{kang2023knowledge,
  title={Knowledge-Augmented Reasoning Distillation for Small Language Models in Knowledge-Intensive Tasks},
  author={Kang, Minki and Lee, Seanie and Baek, Jinheon and Kawaguchi, Kenji and Hwang, Sung Ju},
  journal={arXiv preprint arXiv:2305.18395},
  year={2023}
}
\bibliographystyle{acl_natbib}

\newpage

\appendix
\section*{Appendix}

\begin{table*}[ht]
    \centering
    \begin{tabular}{lcc}
    \toprule
     Method & Model size & Acc  \\
     \midrule
    UL2~\cite{qin2023is_chatgpt_a_general_solver} & 20B & 59.0 \\ 
    StableVicuna INT8~\cite{hartill2023answering} & 13B & 61.7 \\
    GR+RATD~\cite{hartill2023answering} & 440M & 64.2 \\
    KARD~\cite{kang2023knowledge} & 3B & 70.55 \\
    \midrule
    GEEK (ours) & 11B & \textbf{78.17} \\
    \bottomrule
    \end{tabular}
    \caption{More comparison with other baselines.}
    \label{tab:more_results}
\end{table*}

\section{Training Details}
For training the core model, we use the refined annotation of StrategyQA dataset. The original StrategyQA dataset provides human-annotated decomposition questions, fact sentences, and supporting paragraphs from Wikipedia. The decomposition questions break the original open-domain complex question into several simple and explicit sub-questions. The fact sentences contain evidence and background knowledge for solving the main question. The supporting paragraphs are where the fact sentences are grounded, and are one-to-one matched with the corresponding fact sentences. Nonetheless, the fact sentences are not one-to-one mapped to each decomposition question. Therefore, we use GPT4 to refine the annotation, extracting a concise answer for each decomposition question from the set of human-annotated fact sentences. Explicit instruction is given to GPT4 to force the model to use information that is faithful to the human-annotated fact sentences only.

After that, the core model is trained via supervised fine-tuning. Specifically, we use ground-truth annotations (decomposition, paragraphs, and fact sentences) to build a full-solving process for each question. The ground-truth action to be performed also depends on the process. For example, if there are still more ground-truth decomposition questions, the model should select `AddDecomp'. And if all decomposition questions have been visited, the model should select `FinalAnswer'. By doing so, we build multiple input-output pairs simulating each timestep of the solving process, and use these pairs to train the core model in parallel.

\section{More Comparison with Other Baselines}
Besides the results shown in Tab.~\ref{tab:main_results}, we also list more comparison results with other baseline methods, whose backbone models are similar in size to ours. As shown in Tab.~\ref{tab:more_results}, our method significantly surpass the other baselines, with improvement larger than 7.62\%.

\section{Results with Other Backbones}
We also implement GEEK with other backbone models, to verify the generality of our method. In this section, we select the similar-scale LLaMA models for comparison. As shown in Tab.~\ref{tab:more_backbones}, GEEK performs well with all the backbone models. The LLaMA-13B backbone yields similar accuracy as Flan-T5-11B, demonstrating the generality of our GEEK pipeline. However, LLaMA is trained without instruction tuning, and both LLaMA-7B and LLaMA-13B perform slightly worse than Flan-T5-11B. This observation suggests that instruction tuning is helpful for the task of StrategyQA, but a larger model (e.g., LLaMA-13b) can reduce the gap.

\begin{table}[ht]
    \centering
    \begin{tabular}{lc}
    \toprule
     LLM & Acc  \\
     \midrule
    LLaMA-7B & 74.67 \\ 
    LLaMA-13B & 77.73 \\
    Flan-T5-11B & 78.17 \\
    \bottomrule
    \end{tabular}
    \caption{More backbone results.}
    \label{tab:more_backbones}
\end{table}

\section{Prompt Examples}
In this section, we show all the prompts used in GEEK, please also refer to Sec.~\ref{sec:action_space} for more details of the actions. The terms such as $\{Question\_state\}$ represents the placeholder to be substitute by corresponding text values.

\begin{itemize}
    \item System Prompt (at the beginning of every input): 
    \begin{lstlisting}[linewidth=0.9\columnwidth,breaklines=true,language=python, numbers=none]
"Try to solve a multi-step open-domain question. </s>{Question_state}"
    \end{lstlisting}
    \item ActionSelection Prompt1 (begin of each round): 
    \begin{lstlisting}[linewidth=0.9\columnwidth,breaklines=true,language=python, numbers=none]
"Synthesis the information so far, what action should be performed the next? \n[A]: enough evidence to conclude a final answer. \n[B]: ask more sub-questions for necessary knowledge."
    \end{lstlisting}
    \item If final answer (action [A]): 
    \begin{lstlisting}[linewidth=0.9\columnwidth,breaklines=true,language=python, numbers=none]
"Conclude a final answer to the question {Q}. Use the format \"Reasoning steps: [your reasoning steps] Therefore, the final answer is: [yes/no]\""
    \end{lstlisting}
    \item If add decomp (action [B]): 
    \begin{lstlisting}[linewidth=0.9\columnwidth,breaklines=true,language=python, numbers=none]
"Besides the current steps, what more sub-questions need to be asked? Show your steps in the format \"(i) [Q] (sub-question) [A] (pseudo answer)\""
    \end{lstlisting}
    \item ActionSelection Prompt2 (when new decomp is added):
    \begin{lstlisting}[linewidth=0.9\columnwidth,breaklines=true,language=python, numbers=none]
"Does the sub-question {Decomp} involves external evidence to answer? \n[A]: external evidence is needed, call tools. \n[B]: it can be safely answered by logical inference without extra evidence"
    \end{lstlisting}
    \item If call tools (action [A]):
    \begin{lstlisting}[linewidth=0.9\columnwidth,breaklines=true,language=python, numbers=none]
# retriever input
"Question: {Q}, Sub-question: {Decomp}"  

# extractor input
"Based on the following context, answer the question: \"{decomp}\" (A reference but not necessarily correct answer is: \"{pseudo_answer}\")</s>Context: {paragraph}"
    \end{lstlisting}
    \item If self answer (action [B]):
    \begin{lstlisting}[linewidth=0.9\columnwidth,breaklines=true,language=python, numbers=none]
"Based on the sub-questions and facts, use strict logic to answer the sub-question: {Decomp}"
    \end{lstlisting}
\end{itemize}

\section{Error Analysis of GEEK}
We manually analyzed all the error samples in the dev set. The reasons for wrong predictions are categorized into 4 types listed below (with actual examples):
\begin{itemize}
    \item \textbf{Unreasonable decomposition}: the predicted decomposition does not effectively lead to a valid solution to the original answer. For example:
    \begin{itemize}
        \item \textit{Question}: ``Would the author of Little Women have remembered the ratification of the 13th Amendment?"
        \item \textit{Ground truth}: ``(1) When was the 13th Amendment ratified? (2) Who wrote Little Women? [Louisa May Alcott] (3) What years was Louisa May Alcott alive? [1832-1888] (4) Did the ratification of the 13th Amendment occur sometime during 1832-1888?" [Final answer: yes]
        \item \textit{Prediction}: ``(1) When was the 13th Amendment ratified? (2) When was Louisa May Alcott born? [1832] (3) Is 1865 before 1832?" [Final answer: no]
    \end{itemize}
    \item \textbf{Wrong action selection}: the model selects the wrong action to be performed (e.g., conclude final answer too early, incorrectly call retriever, or attempt self-answer)
    \item \textbf{Incorrect facts}: Retriever and extractor output incorrect facts (either irrelavant or factually incorrect). For example:
    \begin{itemize}
        \item \textit{Decomposition question}: ``What are the numbers that are used in the scoring system in table tennis?"
        \item \textit{GT}: ``11 and 21" (for old rules).
        \item \textit{Pred}: ``15 points, 30 points, and 40 points".
    \end{itemize}
    \item \textbf{Logical deduction error}: the model gives the wrong final answer from correct decompositions and facts. For example:
    \begin{itemize}
        \item \textit{Question}: ``Can the Swiss Guard fill the Virginia General Assembly chairs?"
        \item \textit{Facts}: ``There are 140 seats in the Virginia General Assembly. The Swiss Guard has a total of 134 members."
        \item \textit{GT answer}: ``No"
        \item \textit{Pred answer}: ``Yes"
    \end{itemize}
\end{itemize}

Based on the above categories, we also statistic the proportion of errors. The results are shown in Tab.~\ref{tab:err_analysis}. Most of the errors are due to bad decomposition generation and incorrect facts. Because of the nature of implicit QA, generating good decomposition questions is challenging. By observing the error cases, we find that logical reasoning and background knowledge are required and critical to generate high-quality decompositions. Therefore, we hypothesize that GEEK could benefit from a larger backbone LLM, which is believed to have more knowledge absorbed and higher reasoning ability. This is also partially verified by the results in Tab.~\ref{tab:main_results} and Tab.~\ref{tab:more_backbones}.

For the factual errors in generated facts, as also mentioned in the section `Limitation', the extractor's neural processing mechanism makes the hallucination problem inevitable. Based on the observation of error cases, we find that two main reasons result in incorrect facts: (1) the retriever fails to find relevant paragraphs corresponding to the decomposition question, and (2) the extractor outputs summarized fact sentences that are not faithful to the paragraph. For (1), a more powerful retriever (e.g., search engine) or more affluent corpus beyond Wikipedia could help. For (2), faithful QA techniques, such as changing the generative task into an extractive task, might be a remedy. However, constraints by the computation resources, and also considering that this research mainly focuses on implicit question answering, we leave this problem for future works.

\begin{table}[ht]
    \centering
    \begin{tabular}{lc}
    \toprule
     Error type & Percentage  \\
     \midrule
    Unreasonable Decomposition	 & 40\% \\ 
    Wrong action selection & 8\% \\
    Incorrect facts & 54\% \\
    Logical deduction error & 20\% \\
    \bottomrule
    \end{tabular}
    \caption{Error cases and proportion.}
    \label{tab:err_analysis}
\end{table}

\end{document}